%% file: main.tex
\pdfoutput=1

\documentclass[11pt]{article}

\usepackage[preprint]{acl}

\usepackage{times}
\usepackage{latexsym}

\usepackage[T1]{fontenc}

\usepackage[utf8]{inputenc}

\usepackage{microtype}

\usepackage{inconsolata}

\usepackage{graphicx}

\usepackage[textsize=small, disable]{todonotes}
\usepackage{xcolor}
\usepackage{soul}

\usepackage{amsmath} 
\usepackage{amssymb}
\usepackage{xspace}
\newcommand{\proj}{SQFT\xspace}
\usepackage{multirow}
\usepackage{booktabs}
\usepackage{algpseudocode}

\usepackage{tabularray}
\usepackage{amsfonts}
\usepackage{pifont}
\usepackage{arydshln}
\usepackage{algorithm}
\usepackage{algorithmicx}
\usepackage{hyperref}
\usepackage{subcaption}

\newcommand{\xmark}{\ding{55}}%
\usepackage{mathtools}

\usepackage{tikz}
\usepackage{pifont}

\newcommand{\cmark}{\ding{51}}

\title{
\proj: Low-cost Model Adaptation in Low-precision Sparse Foundation Models
}
\author{
 \textbf{J. Pablo Muñoz \textsuperscript{1}}\thanks{
 Co-first authors. 
 },
 \textbf{Jinjie Yuan\textsuperscript{2}}\footnotemark[1],
 \textbf{Nilesh Jain\textsuperscript{1}}
\\
 \textsuperscript{1}Intel Labs,
 \textsuperscript{2}Intel Corporation\\
 \small{
 \{pablo.munoz, jinjie.yuan, nilesh.jain\}@intel.com
 }
}
\begin{document}
\maketitle

\input{content/0_abstract}

\input{content/1_intro}

\input{content/3_methodology}

\input{content/4_experiments}

\input{content/5_conclusion}

\bibliography{main} 
\clearpage
\appendix
\input{content/appendix}

\end{document}

%% file: content/0_abstract.tex
\begin{abstract}
Large pre-trained models (LPMs), such as large language models, have become ubiquitous and are employed in many applications. These models are often adapted to a desired domain or downstream task through a fine-tuning stage. This paper proposes \proj, an end-to-end solution for low-precision sparse parameter-efficient fine-tuning of LPMs, allowing for effective model manipulation in resource-constrained environments. Additionally, an innovative strategy enables the merging of sparse weights with low-rank adapters without losing sparsity and accuracy, overcoming the limitations of previous approaches. \proj also addresses the challenge of having quantized weights and adapters with different numerical precisions, enabling merging in the desired numerical format without sacrificing accuracy. Multiple adaptation scenarios, models, and comprehensive sparsity levels demonstrate the effectiveness of \proj. Models and code are available at \href{https://github.com/IntelLabs/Hardware-Aware-Automated-Machine-Learning}{https://github.com/IntelLabs/Hardware-Aware-Automated-Machine-Learning}.
\end{abstract}

%% file: content/1_intro.tex
\section{Introduction}

Despite several limitations, such as hallucinations and a significant computational footprint, large pre-trained, foundation, or frontier models have become integral to numerous applications, including language understanding and code generation. These models are trained with extensive corpora on thousands of graphics processing units (GPUs), resulting in outstanding zero-shot performance across various tasks and datasets. However, it is frequently the case that they must be adapted to improve their performance on new tasks or data.  

\begin{figure}
  \centering
  \includegraphics[width=\linewidth]{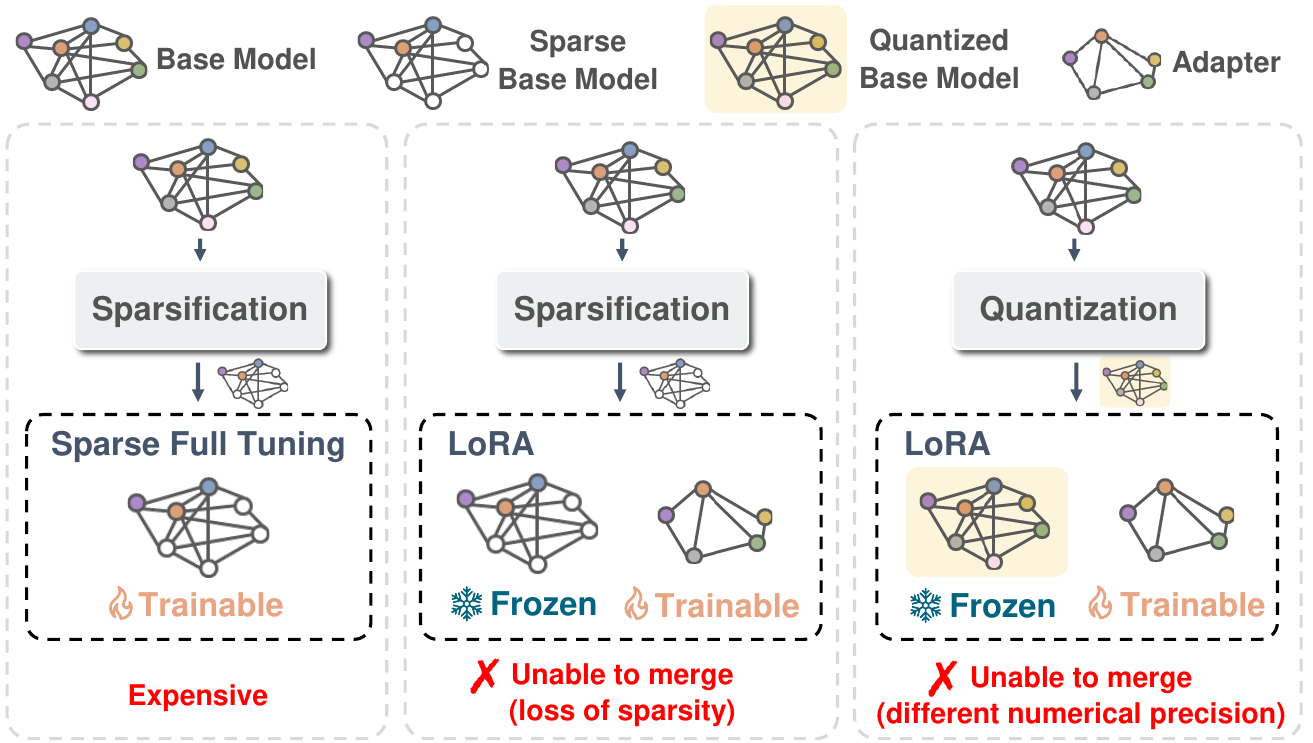}
  \caption{Limitations of existing approaches for fine-tuning sparse and quantized models. Full fine-tuning is expensive. Low-rank adapters (LoRA) for Parameter-efficient Fine-tuning (PEFT) on sparse or quantized models cannot easily merge with the compressed weights due to loss of previously induced sparsity or different numerical precision.
  }
\label{fig:limitation}
\end{figure}

Low-rank adapters (LoRA) \cite{hu2022lora} have demonstrated their effectiveness in model adaptation. However, when LoRA is combined with model compression techniques, e.g., sparsity or quantization, several challenges prevent merging these adapters into a single compressed and fine-tuned model, as illustrated in Figure \ref{fig:limitation}. These challenges stem from two primary reasons: \textbf{i)} merging dense adapters causes the loss of sparsity in the base model, and \textbf{ii)} adapter merging cannot be achieved due to different numerical precisions.

\begin{figure*}[ht]
    \includegraphics[width=\textwidth]{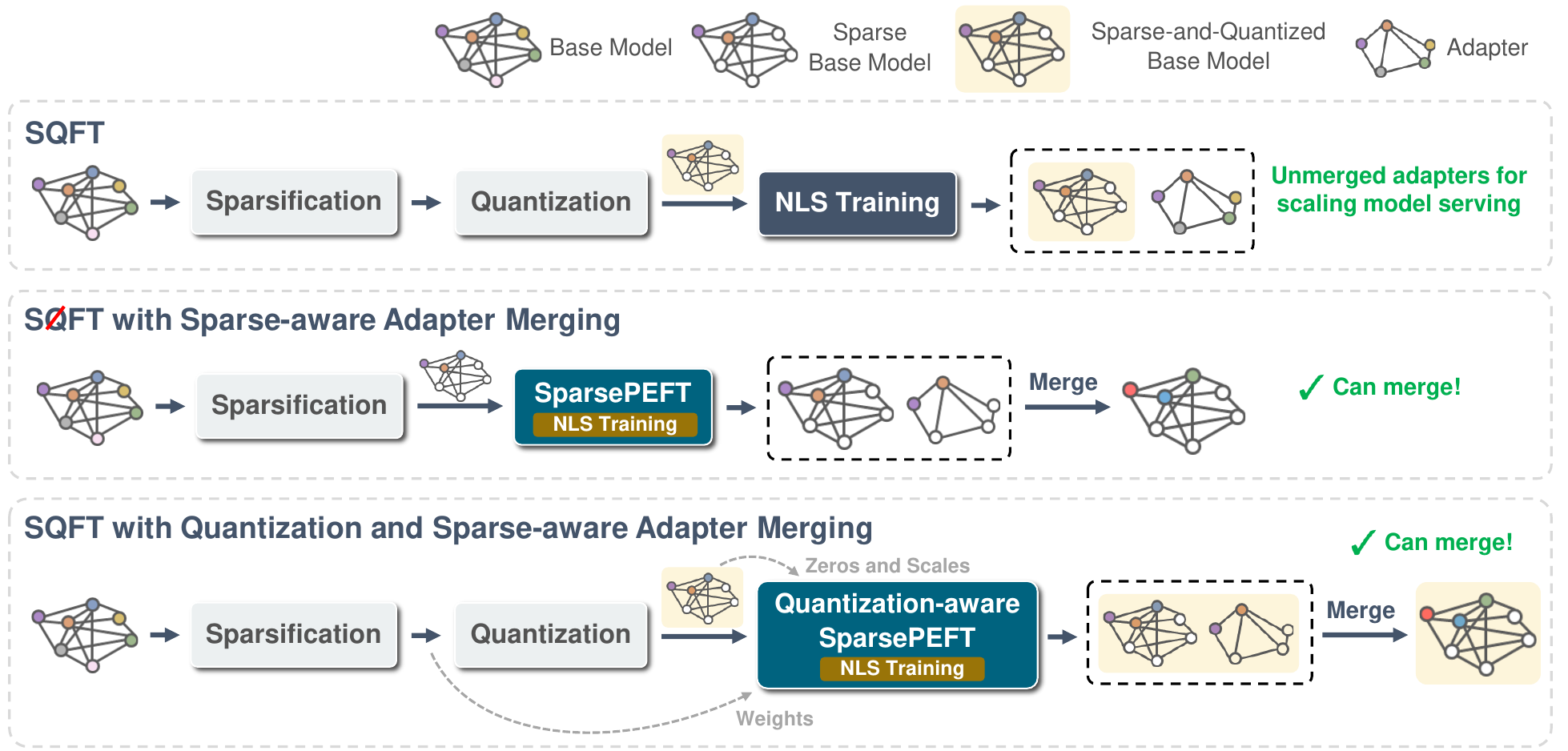}
    \caption{\proj Overview. Several pipeline configurations can be utilized to efficiently fine-tune large models while addressing several limitations of existing approaches.  
}\label{fig:overall_pipeline}    
\end{figure*}

This paper introduces \textbf{\proj}, an end-to-end compression and model adaptation solution for large pre-trained models (LPMs) that alleviates the limitations above. \proj is designed to sparsify, quantize, and fine-tune large models and can instantiate efficient pipelines that streamline compression techniques. 
Within the \proj framework, we propose \textbf{Sparse} \textbf{P}arameter-\textbf{E}fficient \textbf{F}ine-\textbf{T}uning (\textbf{SparsePEFT}), a strategy to address the adapter merging problem for sparse and quantized models, resulting in more effective high-performing models.
Furthermore, \proj also benefits from weight-sharing techniques applied to traditional parameter-efficient fine-tuning (PEFT) techniques and incorporates insights from state-of-the-art compression techniques. Throughout this paper, we discuss the following contributions:  

\begin{enumerate}
    \item An end-to-end model adaptation solution, \proj, designed for efficient low-cost configurable pipelines tailored for large pre-trained models with low numerical precision and sparsity. 
    \item SparsePEFT, a component of \proj, addresses several limitations in existing parameter-efficient fine-tuning approaches for sparse and quantized models, including the reduction in the cost of fine-tuning, the effective merging of adapters into the sparse model without the loss of sparsity, and the effective merging of components that operate in different numerical precision. 
    \item Extensive experiments demonstrate the effectiveness of \proj across different foundation models, sparsity levels, and adaptation scenarios.
\end{enumerate}

This paper is organized as follows: Section \ref{sec:method} describes the stages in the proposed end-to-end solution, \proj. Section \ref{seq:experiments} discusses \proj's evaluation, and we finalize with some concluding remarks in Section \ref{sec:conclusion}. Due to page limits, we include a Related Work section and additional results in the Appendix.

%% file: content/3_methodology.tex
\section{Methodology}
\label{sec:method}

\proj fine-tunes large pre-trained models (LPMs) in an efficient multi-stage approach that includes (1) Sparsification, with an optional reduction in the numerical precision, i.e., Quantization, (2) Fine-tuning with Neural Low-rank Adapter Search (NLS), (3) Sparse Parameter-Efficient Fine-Tuning (SparsePEFT) with optional (4) Quantization-awareness. Figure \ref{fig:overall_pipeline} illustrates the alternative model compression and adaptation pipelines that were explored. In the following sections, we discuss the details of each stage and the benefits of accelerating inference and model serving.

\subsection{Sparsification and Quantization Stage} 
As shown in Figure \ref{fig:overall_pipeline}, at the beginning of all possible pipeline configurations, \proj employs an effective method to induce sparsity in the model. For a given weight matrix $\boldsymbol{W} \in \mathbb{R}^{m\times n}$, with entries $w_{i, j}$ s.t. $\boldsymbol{W} = (w_{i,j}), 1 \leq i \leq m, 1 \leq j \leq n$, an arbitrary scoring function, $\Psi$, is assigned to the proposed solution. This function determines the relative importance of $w_{i,j}$ compared to the other weights in $\boldsymbol{W}$. $\Psi$ can be formulated in various ways. For instance, $\Psi(\boldsymbol{W}) = |\boldsymbol W| \cdot \|\boldsymbol X\|_2$, where $\boldsymbol X$ represents sampled feature input activations, as proposed by \citet{sun2023wanda}. However, it is important to highlight that the proposed end-to-end model fine-tuning solution, \proj, can utilize any other scoring function. Leveraging the scores from $\Psi$ and a desired level of sparsity, $s$, we derive the sparsified weight, denoted as $\boldsymbol{W}^p$, with a sparsity pattern $S\{\boldsymbol{W}^p\} = \{(i, j) \mid \boldsymbol{W}^p_{i, j} \not= 0, 1 \leq i \leq m, 1 \leq j \leq n\}$, s.t. $\lvert S\{\boldsymbol{W}^p\}\rvert \leq \lvert S\{\boldsymbol{W}\}\rvert$. 

It has been demonstrated that LPMs can tolerate higher sparsity levels compared with the previous generations of smaller transformer-based models \cite{frantar-sparsegpt}. Our experiments confirm these observations (Section \ref{seq:experiments}).
SQFT's evaluations use Wanda \cite{sun2023wanda} to measure the importance and replace the least important base model's weights with zeros. Once sparsity has been induced in the pre-trained weights, $\boldsymbol{W}^p$, we might enable an optional reduction in their numerical precision. Given the sparse weights, \proj applies layer-wise one-shot quantization 
\cite{nagel_up_or_down, frantar-gptq, pmlr-v119-wang20c_towards_acc, frantar2022obc}. 
Utilizing a selection from state-of-the-art post-training quantization approaches, \proj identifies the low-precision sparse weights, denoted as $\widehat{\boldsymbol{W}}^p$, that given an input $\boldsymbol{X}$, minimize $argmin_{\widehat{\boldsymbol{W}}^p} \lvert\lvert \boldsymbol{W}^p\boldsymbol{X} - \boldsymbol{\widehat{W}}^p\boldsymbol{X}\rvert\rvert^2_2$.
In SQFT's evaluation (Section \ref{seq:experiments}), we use GPTQ \cite{frantar-gptq}, but other similar approaches can be used to obtain the quantized weights.

Reducing the numerical precision and inducing sparsity on weights frequently decrease the model's accuracy, requiring fine-tuning to improve performance.  

\subsection{Fine-tuning with Neural Low-rank Adapter Search (NLS)}

Given the sparse quantized weights, $\boldsymbol{\widehat{W}}^p$, \proj recovers any drops in accuracy induced by the compression schema and fine-tunes these weights for a specific downstream task. 
As shown in Figure \ref{fig:overall_pipeline}, \proj employs Neural Low-rank Adapter Search (NLS) \cite{shears} instead of vanilla Low-rank Adapters (LoRA) \cite{hu2022lora}, and fine-tunes sparse and quantized models. To justify using NLS, traditional LoRA adapters require assigning the values for several hyperparameters, including their rank $r$, and the subset of modules where these adapters will be placed. Determining these hyperparameters can be a challenging endeavor. To alleviate this limitation, \proj extends NLS' weight-sharing techniques to facilitate the discovery of optimal adapter configurations from a space of elastic adapter configurations. In other words, instead of having a fixed value for the rank, $r$, we enable elastic configurations, $C = [c_1, \ldots, c_n]$, s.t., $r \leftarrow c_i$ depending on the activation of the corresponding sub-adapter.

\subsection{SparsePEFT}

\begin{figure}
  \centering
\includegraphics[width=\linewidth]{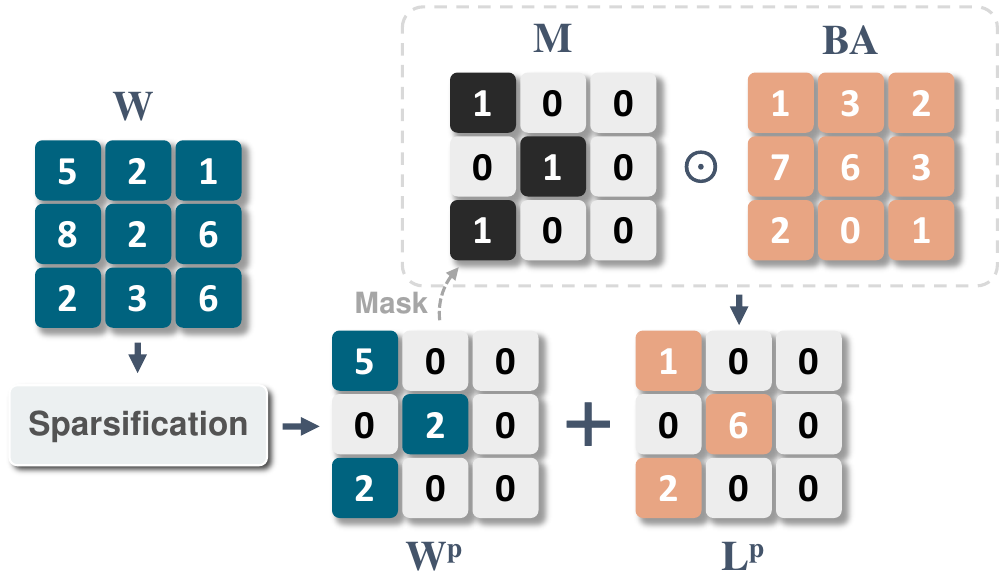}
  \caption{Sparse Parameter-efficient Fine-tuning (SparsePEFT). A binary mask is obtained from the sparsified weights and applied to the adapters, allowing for the later merge without loss of sparsity.
  }
\label{fig:sparsepeft}
\end{figure}

Fine-tuning the sparse quantized model with elastic adapters effectively improves the model's performance on a downstream task. However, as illustrated in the middle and right part of Figure \ref{fig:limitation}, a challenge arises when dealing with sparse or quantized weights and dense adapter weights: merging them will i) result in the loss of sparsity on the model's weights or ii) be unable to merge due to different numerical precisions. 
Aiming to address the first limitation, we propose an effective strategy, Sparse Parameter-Efficient Fine-Tuning (SparsePEFT), to make adapters sparsity-aware. 
As depicted in Figure \ref{fig:sparsepeft}, SparsePEFT applies a binary mask $\boldsymbol{M}$ derived from the initial sparsification of $\boldsymbol{W}$. This mask is used to sparsify the adapters matrix (denoted as $\boldsymbol{BA}$) into $\boldsymbol{L}^p$. The process can be formulated as: 
\begin{equation}
\boldsymbol{L}^p = (\boldsymbol{BA}) \odot \boldsymbol{M},
\end{equation}
which is activated during the fine-tuning process for sparsity awareness. 
SparsePEFT enables the merging of the sparsified weights $\boldsymbol{W}^p$ and the adapter weight $\boldsymbol{L}^p$ without sacrificing the sparsity induced early in the compression pipeline as follows, 
\begin{equation}
    \boldsymbol{W}^p \leftarrow \boldsymbol{W}^p + \boldsymbol{L}^p.
\end{equation}
In addition to preserving sparsity, SparsePEFT demonstrates comparable (even better) accuracy compared to fine-tuning with dense adapters. Extensive experimental findings substantiate the advantages of SparsePEFT, as detailed in Section \ref{seq:experiments}.  
  
Although SparsePEFT can effectively preserve the model's sparsity, it presents additional challenges when merging with quantized models, the second limitation we discussed before, which is primarily attributed to the need for the adapter and pre-trained weights to possess identical numerical precision. 
In the following subsection, we explore a pipeline variation for \proj that facilitates the integration of sparse quantized weights. This approach aims to address both challenges mentioned above while improving the overall efficiency of the resulting model.

\subsection{Quantization-aware SparsePEFT}

Building upon the concept of SparsePEFT, we propose Quantization-aware SparsePEFT (QA-SparsePEFT), an extension of SparsePEFT for sparse quantized models. QA-SparsePEFT integrates quantization awareness into SparsePEFT. 
In most common quantization schemes, the zero point and scales for the target quantized tensor are determined during the quantization process.
Within the QA-SparsePEFT stage, the zeros and scales of the sparse quantized weights, $\widehat{\boldsymbol{W}}^p$, of the based model are shared with the adapter. 
The elastic adapters can then be quantized smoothly with the shared fixed zeros and scales, enabling quantization-aware fine-tuning.
Formally, given the sparsified pre-trained weight $\boldsymbol{W}^p$, sparsified adapter weight $\boldsymbol{L}^p$ obtained from SparsePEFT, zeros $\boldsymbol{z}$ and scales $\boldsymbol{s}$ from the quantization of $\boldsymbol{W}^p$, the quantization process in the proposed QA-SparsePEFT can be formulated as:
\begin{equation}
\small
   \widehat{\boldsymbol{W}}^p_m = \text{clamp}\left( \text{round}\left( \frac{\boldsymbol{W^p+L^p}}{\boldsymbol{s}} \right) + \boldsymbol{z},0,Q_p\right),
\end{equation}
where $\widehat{\boldsymbol{W}}^p_m$ denotes the sparse quantized (merged) weight and $Q_p=2^{n-1}-1$ (n represents the bit-width of the quantized values). Dequantization is the inverse as follows:
\begin{equation}
\small
   \tilde{\boldsymbol{W}}^p_m = \boldsymbol{s} \left( \widehat{\boldsymbol{W}}^p_m - \boldsymbol{z} \right),
\end{equation}
which applies $\boldsymbol{z}$ and $\boldsymbol{s}$ to approximate $\boldsymbol{W}^p_m$.
Through QA-SparsePEFT, we can obtain the fine-tuned, sparse, low-precision resulting model.
Moreover, \proj with QA-SparsePEFT can run the NLS stage using this schema, which allows us to merge the adapters as soon as an optimal configuration has been discovered.

\subsection{Model Serving and Inference Acceleration}

Accelerating model serving and inference through sparsification and quantization techniques has shown significant efficacy across various hardware platforms and kernels, demonstrating remarkable speedups. However, adding adapter modules for PEFT with a sparse or quantized model (as shown in Figure \ref{fig:limitation}) introduces computational overhead during inference due to their non-mergeability.
SparsePEFT and QA-SparsePEFT allow adapters to be merged into the sparse and quantized model, which can reduce adapters' redundancy and computational overhead, leading to more streamlined inference processes.
Moreover, quantization techniques further enhance acceleration by reducing the model size and computational complexity, but balancing the trade-off between acceleration and maintaining competitive accuracy is essential.

In summary, \proj and its SparsePEFT strategy bring the benefits of adapter merging and maintaining accuracy on sparse or quantization scenarios. The choice between the sparsity level and whether to apply quantization depends on the specific deployment scenario (e.g., task requirements and resource constraints), including the trade-off between model performance, inference speed, and memory efficiency. 
In the next section, we will delve into further empirical studies to fully understand the strengths and weaknesses of each approach in different settings.

%% file: content/4_experiments.tex
\section{Experimental Results}
\label{seq:experiments}

\begin{table*}[hbt]
\setlength{\belowcaptionskip}{0.5pt}
\setlength{\tabcolsep}{8pt}
\caption{Results from adapting \textbf{Llama-3-8B} and \textbf{Mistral-7B-v0.3} to GSM8K. The criterion for mergeable is that there should be no loss in either accuracy or sparsity before and after merging. The evaluation used the default configuration for \textit{lm-eval-harness} \cite{eval-harness} (5-shot).
}
\label{tab:gsm8k_main_sparse}
\scriptsize 
\centering
\renewcommand\arraystretch{1.2}
\begin{tabular}{lclccc}
\toprule
\multirow{2}{*}{\textbf{Model}} & \multirow{2}{*}{\textbf{Sparsity}}  & \multirow{2}{*}{\textbf{Method}} & \multirow{2}{*}{\textbf{Mergeable}} & \textbf{Final Precision}  & \textbf{GSM8K Test} \\ 
&  & & & \textbf{(Base + Adapter / Base)}   & \textbf{Accuracy(\%)}   \\
\midrule

\multirow{11}{*}{\textbf{Llama-3-8B}} & 0\%  & w/o tune &  -  & FP16  & 50.0  \\
\cline{2-6}
& \multirow{10}{*}{50\%} &  \multicolumn{4}{c}{\textit{w/o Quantization}} \\
\cdashline{3-6}
& & w/o tune &  -  & FP16  & 12.5   \\
& & LoRA        & \xmark & FP16 + FP16 & 50.6  \\
& & Shears        & \xmark & FP16 + FP16 & 52.2  \\
& & \proj + SparsePEFT (Ours) & \cmark & FP16  &\textbf{52.5} \\
\cdashline{3-6}
& & \multicolumn{4}{c}{\textit{Quantization}} \\
\cdashline{3-6}
& & w/o tune &  -  & \textbf{INT4}  & 7.0   \\
& & GPTQ + LoRA        & \xmark & INT4 + FP16 & 48.9  \\
& & \proj (Ours)       & \xmark & INT4 + FP16   & 50.0 \\
& & \proj + QA-SparsePEFT (Ours) & \cmark & \textbf{INT4}   & \textbf{50.2}  \\

\midrule

\multirow{11}{*}{\textbf{Mistral-7B-v0.3}} & 0\%  & w/o tune &  -  & FP16  & 36.0  \\
\cline{2-6}
& \multirow{10}{*}{50\%} &  \multicolumn{4}{c}{\textit{w/o Quantization}} \\
\cdashline{3-6}
& & w/o tune &  -  & FP16  & 17.2   \\
& & LoRA        & \xmark & FP16 + FP16 & 44.1 \\
& & Shears        & \xmark & FP16 + FP16 & 45.1  \\
& & \proj + SparsePEFT (Ours) & \cmark & FP16  &\textbf{50.1} \\
\cdashline{3-6}
& & \multicolumn{4}{c}{\textit{Quantization}} \\
\cdashline{3-6}
& & w/o tune &  -  & \textbf{INT4}  & 16.0   \\
& & GPTQ + LoRA        & \xmark & INT4 + FP16 & 44.0  \\
& & \proj (Ours)       & \xmark & INT4 + FP16   & \textbf{44.5} \\
& & \proj + QA-SparsePEFT (Ours) & \cmark & \textbf{INT4}   & 44.0  \\

\bottomrule
\end{tabular}
\end{table*}

We evaluate \proj on several state-of-the-art large pre-trained models and datasets. Next, we discuss the setup for our experiments. 

\subsection{Setup}
\label{sec:ex_setup}

\paragraph{Models} \proj is evaluated on three state-of-the-art models, including 
Llama-3-8B\footnote{https://huggingface.co/meta-llama/Meta-Llama-3-8B}, Mistral-7B-v0.3\footnote{https://huggingface.co/mistralai/Mistral-7B-v0.3} and Phi-3-Mini-4K-Instruct\footnote{https://huggingface.co/microsoft/Phi-3-mini-4k-instruct}.  
To study it more comprehensively, we aim to explore \proj across different models, scales, and settings.

\paragraph{Datasets and Downstream Tasks} Aligned with other works in the LPMs compression and fine-tuning spaces, \proj is validated on three experimental settings: 
\textbf{1)} Grade School Math 8K (GSM8K) \cite{cobbe2021training_GSM8K}, \textbf{2)} Math reasoning with instruction tuning (following LLM-Adapters \cite{hu2023llm_adapters}), including 3 math reasoning datasets: GSM8K, 
Math Word Problems (MAWPS)
\cite{koncel-kedziorski-etal-2016-mawps}, Simple Variations on Arithmetic Math word Problems (SVAMP)
\cite{patel-etal-2021-nlp-svamp}, and \textbf{3)} Commonsense reasoning datasets: Boolean Questions (BoolQ)
\cite{clark2019boolq}, Physical Interaction: Question Answering (PIQA)
\cite{Bisk2020_piqa}, 
HellaSwag
\cite{zellers2019hellaswag}, Large-scale Winograd Schema Challenge (WinoGrande)
\cite{winogrande}, AI2 Reasoning Challenges (Arc-e, Arc-c)
\cite{Clark2018ThinkYH_arc}, and 
Open Book Question Answering (OBQA) \cite{Mihaylov2018CanAS_obqa}.

\paragraph{Evaluation Settings} The evaluations of our experiments are conducted utilizing \emph{lm-eval-harness} \cite{eval-harness} in both setting 1 and 3 while following the evaluation from LLM-Adapters in setting 2. We present a comparative analysis of the results obtained from our various pipelines and also compare with vanilla LoRA \cite{hu2022lora}, Shears \cite{shears} (a parameter-efficient fine-tuning method for sparse models), and GPTQ + LoRA. For fair comparison, all methods are run in the same environment and with the same configuration.
\proj employs the implementation of Wanda \cite{sun2023wanda} as default method for sparsification, and GPTQ in Huggingface \footnote{https://huggingface.co/blog/gptq-integration} for quantizing the LPMs and adapters. The hyperparameters used in our experiments are detailed in the Appendix. 

\paragraph{Reference Configuration} Unless stated in the results, we report a reference configuration for \proj. This configuration is obtained utilizing the heuristic proposed in \citet{lonas2024}. The heuristic is intuitive and straightforward, activating the configuration with the median of each set of elastic values per module. Spending additional cycles to search the space of configurations might yield even more competitive results, presented in Table \ref{tab:hillclimbing}. Next, we discuss experimental results and studies conducted using \proj. 

\begin{table*}[hbt]
\setlength{\belowcaptionskip}{0.5pt}
\setlength{\tabcolsep}{2pt}
\caption{Results from adapting \textbf{Mistral-7B-v0.3} and \textbf{Phi-3-Mini-4K-Instruct} with math instruction tuning. \emph{Mergeable} means that merging the dense adapters with the sparse weights is possible without losing the induced sparsity levels or affecting the desired low numerical precision.
}
\label{tab:math_main}
\scriptsize 
\centering
\renewcommand\arraystretch{1.2}
\begin{tabular}{lclcccccc}
\toprule
\multirow{2}{*}{\textbf{Model}} & \multirow{2}{*}{\textbf{Sparsity}}  & \multirow{2}{*}{\textbf{Method}} & \multirow{2}{*}{\textbf{Mergeable}} & \textbf{Final Precision}  & \multicolumn{3}{c}{\textbf{Datasets | Accuracy(\%)}} & \multirow{2}{*}{\textbf{Average}} \\ 
&  & & & \textbf{(Base + Adapter / Base)}   & \textbf{GSM8K} & \textbf{MAWPS} & \textbf{SVAMP} &   \\

\midrule

\multirow{11}{*}{\textbf{Mistral-7B-v0.3}} & 0\%  & w/o tune &  -  & FP16  & - & - & - & - \\
\cline{2-9}
& \multirow{10}{*}{50\%} &  \multicolumn{7}{c}{\textit{w/o Quantization}} \\
\cdashline{3-9}
& & w/o tune &  -  & FP16 & - & - & - & -  \\
& & LoRA        & \xmark & FP16 + FP16 & 53.8 & 85.7 & 58.2 & 65.9 \\
& & Shears        & \xmark & FP16 + FP16 & 53.0 &  87.4 &  61.7 & 67.4 \\
& & \proj + SparsePEFT (Ours) & \cmark & FP16  & 55.3 & 87.4 & 59.8 & \textbf{67.5} \\
\cdashline{3-9}
& & \multicolumn{7}{c}{\textit{Quantization}} \\
\cdashline{3-9}
& & w/o tune &  -  & \textbf{INT4}  & - & - & - & -  \\
& & GPTQ + LoRA        & \xmark & INT4 + FP16 & 51.4 & 87.4 & 60.3 & 66.4 \\
& & \proj (Ours)       & \xmark & INT4 + FP16   & 51.3 & 87.0 & 62.8 & 67.0 \\
& & \proj + QA-SparsePEFT (Ours) & \cmark & \textbf{INT4}   & 54.1 & 88.2 & 59.1 & \textbf{67.2} \\

\midrule

\multirow{11}{*}{\textbf{Phi-3-Mini-4K-Instruct}} & 0\%  & w/o tune &  -  & FP16  & 64.7 & 84.5 & 85.4 & 78.2 \\
\cline{2-9}
& \multirow{10}{*}{50\%} &  \multicolumn{7}{c}{\textit{w/o Quantization}} \\
\cdashline{3-9}
& & w/o tune &  -  & FP16 & 38.9 & 64.7 & 66.8 & 56.8 \\
& & LoRA        & \xmark & FP16 + FP16 & 62.5 & 90.3 & 77.8 & 76.9 \\
& & Shears        & \xmark & FP16 + FP16 & 62.3 &  90.8 &  76.1 & 76.4 \\
& & \proj + SparsePEFT (Ours) & \cmark & FP16  & 61.9 & 91.2 & 78.7 & \textbf{77.3} \\
\cdashline{3-9}
& & \multicolumn{7}{c}{\textit{Quantization}} \\
\cdashline{3-9}
& & w/o tune &  -  & \textbf{INT4}  & 33.4 & 56.7 & 64.2 & 51.4 \\
& & GPTQ + LoRA        & \xmark & INT4 + FP16 & 60.3 & 89.5 & 74.8 & 74.9 \\
& & \proj (Ours)       & \xmark & INT4 + FP16   & 60.3 & 90.8 & 75.6 & \textbf{75.5} \\
& & \proj + QA-SparsePEFT (Ours) & \cmark & \textbf{INT4}   & 61.8 & 88.7 & 75.5 & 75.3 \\

\bottomrule
\end{tabular}
\end{table*}
 
\subsection{Main Results}
\subsubsection{Fine-tuning on GSM8K}

We begin our evaluation with Llama-3-8B and Mistral-7B-v0.3, assessing their accuracy in a dense mode and after inducing 50\% sparsity without fine-tuning on the GSM8K dataset. Subsequently, we execute various pipelines of \proj.
As described in Table \ref{tab:gsm8k_main_sparse}, for Llama-3-8B at the 50\% sparsity level, \proj recovers the model's accuracy from 12.5\% to 52.5\% without employing quantization, while allowing for the merging of adapters without sacrificing sparsity (SparsePEFT) and incorporating quantization into the pipeline results in a minor drop in accuracy to 50.2\% when enabling the adjustment to merge adapters (QA-SparsePEFT). 

More importantly, \proj with SparsePEFT and QA-SparsePEFT exhibit comparable performance to their corresponding non-mergeable approaches. This behavior is particularly evident in the non-quantized experimental setup for the Mistral-7B-v0.3 model, where SQFT + SparsePEFT (50.1\%) significantly outperforms its two baselines, LoRA (44.1\%) and Shears (45.1\%). These results suggest that \proj with SparsePEFT (QA-SparsePEFT) effectively addresses the limitation of the merging problem encountered when fine-tuning adapters into sparse models (or sparse and quantized models) without any degradation in accuracy.
Furthermore, the comparison between LoRA and \proj with SparsePEFT (or Shears), and between GPTQ + LoRA and \proj, highlights the superior performance of NLS (elastic rank) compared with LoRA (fixed rank).
We explore the performance of a broader range of sparsity levels and conduct more detailed ablation experiments in this experimental setting, which can be found in Sections \ref{sec:broader_range_of_sparsity_levels} and \ref{sec:ablation_Studies}, respectively. The Appendix includes ablation experiments without sparsity and only utilizing SQFT to fine-tune quantized models.

\subsubsection{Math Reasoning with Instruction Tuning}

In addition to fine-tuning on GSM8K, we also investigated the performance of \proj with Mistral-v0.3 and Phi-3. Since the Phi-3-series models released by Microsoft are the best-suited instruction models for a chat prompt, we evaluate \proj on three math reasoning datasets for instruction tuning.  
Table \ref{tab:math_main} presents the test accuracy for our approaches and baselines. 
Interestingly, in the full-precision mode (\textit{w/o Quantization}), our proposed SparsePEFT not only achieves the highest average accuracy (67.5\% for Mistral-v0.3 and 77.3\% for Phi-3) compared to other approaches but also uniquely allows for the merging of adapters and sparse weights without any loss of sparsity. This result is achieved without needing an expensive search and by utilizing the heuristic detailed in Section \ref{sec:ex_setup}.
In the quantization mode, the accuracy of \proj + QA-SparsePEFT (mergeable) is comparable to the non-mergeable approaches (67.2\% vs. 66.4\%/67.0\% and 75.3\% vs. 74.9\%/75.5\%). This result suggests a need to balance the trade-off between accuracy and efficiency. Fortunately, \proj + QA-SparsePEFT results in a merged fine-tuned quantized model, eliminating the overhead associated with dense adapters.

\subsubsection{Fine-tuning on Commonsense Reasoning}

\begin{table*}[hbt]
\setlength{\belowcaptionskip}{0.5pt}
\setlength{\tabcolsep}{0.1pt}
\caption{Results from adapting \textbf{Phi-3-Mini-4K-Instruct} with commonsense reasoning. \proj obtains competitive fine-tuned models with an additional benefit over Shears and LoRA applied to low-precision weights, i.e., \proj's adapters can be efficiently merged into the weights without any loss of precision or accuracy. We are reporting a reference submodel for \proj obtained the heuristic detailed in \ref{sec:ex_setup}, which means that, as shown in Table \ref{tab:hillclimbing}, with an additional cost, \proj can discover submodels with even higher performance. }
\label{tab:cs_main}
\scriptsize 
\centering
\renewcommand\arraystretch{1.5}
\begin{tabular}{lclcccccccccc}
\toprule
\multirow{2}{*}{\textbf{Model}} & \multirow{2}{*}{\textbf{Sparsity}}  & \multirow{2}{*}{\textbf{Method}} & \multirow{2}{*}{\textbf{Mergeable}} & \textbf{Final Precision}  & \multicolumn{7}{c}{\textbf{Datasets | Accuracy(\%)}} & \multirow{2}{*}{\textbf{Average}} \\ 
&  & & & \textbf{(Base + Adapter / Base)}   & \textbf{BoolQ} & \textbf{PIQA} & \textbf{HellaS} & \textbf{WinoG}& \textbf{Arc-e}& \textbf{Arc-c}& \textbf{OBQA} &   \\
\midrule

\multirow{11}{*}{\textbf{Phi-3-Mini-4K-Instruct}} & 0\%  & w/o tune &  -  & FP16  & 86.1 & 80.3 & 78.5 & 73.7 & 83.2 & 57.5 & 46.8 & 72.3 \\
\cline{2-13}
& \multirow{10}{*}{50\%} &  \multicolumn{11}{c}{\textit{w/o Quantization}} \\
\cdashline{3-13}
& & w/o tune &  -  & FP16 & 82.5 & 75.9 & 69.9 & 69.1 & 76.9 & 50.9 & 43.4 & 66.9 \\
& & LoRA        & \xmark & FP16 + FP16 & 85.6 & 79.1 & 75.8 & 71.5 & 79.6 & 53.2 & 49.4 & 70.6 \\
& & Shears        & \xmark & FP16 + FP16 & 85.2 & 78.9 & 75.7 & 72.6 & 80.1 & 53.3 & 50.4 & \textbf{70.9} \\
& & \proj + SparsePEFT (Ours) & \cmark & FP16  & 84.0 & 78.8 & 75.5 & 72.1 & 80.1 & 53.5 & 48.6 & 70.4 \\
\cdashline{3-13}
& & \multicolumn{11}{c}{\textit{Quantization}} \\
\cdashline{3-13}
& & w/o tune &  -  & \textbf{INT4}  & 81.4 & 75.2 & 68.5 & 68.2 & 75.9 & 50.3 & 40.2 & 65.7 \\
& & GPTQ + LoRA        & \xmark & INT4 + FP16   & 85.3 & 79.1 & 75.3 & 72.5 & 79.5 & 54.6 & 47.2 & 70.5 \\
& & \proj (Ours)        & \xmark & INT4 + FP16   & 85.1 & 79.0 & 75.4 & 71.2 & 79.6 & 54.1 & 48.8 & 70.5 \\
& & \proj + QA-SparsePEFT (Ours) & \cmark & \textbf{INT4}   & 83.7 & 80.1 & 74.1 & 73.6 & 80.1 & 55.1 & 48.2 & \textbf{70.7} \\

\bottomrule
\end{tabular}
\end{table*}

\begin{table*}[hbt]
\setlength{\belowcaptionskip}{0.5pt}
\setlength{\tabcolsep}{2pt}
\caption{ Hill-climbing searching results for \textbf{Phi-3-Mini-4K-Instruct} with the commonsense reasoning dataset.
}
\label{tab:hillclimbing}
\tiny 
\centering
\renewcommand\arraystretch{1.2}
\begin{tabular}{lcllcccccccccccc}
\toprule
\multirow{2}{*}{\textbf{Model}} & \multirow{2}{*}{\textbf{Sparsity}}  & \multirow{2}{*}{\textbf{Method}} & \multirow{2}{*}{\textbf{Sub-Adapter}} & \multicolumn{4}{c}{\textbf{Validation Datasets | Accuracy(\%)}} & \multicolumn{8}{c}{\textbf{Test Datasets | Accuracy(\%)}} \\ 
&  & & & \textbf{Arc-e}& \textbf{Arc-c}& \textbf{OBQA} &  \textbf{Average} & \textbf{BoolQ} & \textbf{PIQA} & \textbf{HellaS} & \textbf{WinoG}& \textbf{Arc-e}& \textbf{Arc-c}& \textbf{OBQA} &  \textbf{Average} \\
\midrule

\multirow{4}{*}{\textbf{Phi-3-Mini-4K-Instruct}} & \multirow{4}{*}{50\%} & \multirow{2}{*}{\proj + SparsePEFT} & Heuristic & 79.3 & 50.8 & 47.4 & 59.2 & 84.0 & 78.8 & \textbf{75.5} & \textbf{72.1} & 80.1 & 53.5 & 48.6 & 70.4 \\
 &  &  & Hill-climbing & \textbf{80.2} & \textbf{51.8}  & \textbf{47.6} & \textbf{59.9} & \textbf{84.3}	&\textbf{78.9}&	75.4	&72.0&	80.1&	\textbf{54.3}&	\textbf{49.4}	&\textbf{70.6}  \\
\cdashline{3-16}
& & \multirow{2}{*}{\proj + QA-SparsePEFT} & Heuristic & 80.0 & 51.5 & 45.4 & 59.0 & \textbf{83.7} & \textbf{80.1} &\textbf{ 74.1} & 73.6 & 80.1 & 55.1 & 48.2 & 70.7 \\
& & & Hill-climbing & \textbf{80.4} & \textbf{53.5} & \textbf{46.2} & \textbf{60.0} & 83.6&	79.7	&\textbf{74.1}&	\textbf{73.7}&	80.1&	\textbf{56.2}&	\textbf{48.8}&	\textbf{70.9} \\

\bottomrule
\end{tabular}
\end{table*}

Besides the mathematical domain of the first two experimental settings, we also explore \proj in other areas, e.g., commonsense reasoning.
We apply \proj to fine-tuning the Phi-3 model on a set of unified commonsense training datasets with 83K samples for fine-tuning from BoolQ, PIQA, HellaSwag, WinoGrande, Arc-e, Arc-c, and OBQA. Table \ref{tab:cs_main} compares the test accuracy of the evaluated approaches. \proj obtains a competitive configuration with Shears, LoRA, and GPTQ + LoRA. However, \proj has the additional benefit of allowing for the merging without losing the previously induced sparsity, both in full-precision and quantized modes.
It is worth noting that \proj with QA-SparsePEFT shows super competitiveness here, i.e., the most efficient model with high accuracy (among all full-precision and quantized cases).

\begin{figure}
  \centering
  \includegraphics[width=\linewidth, height=0.27\textheight]{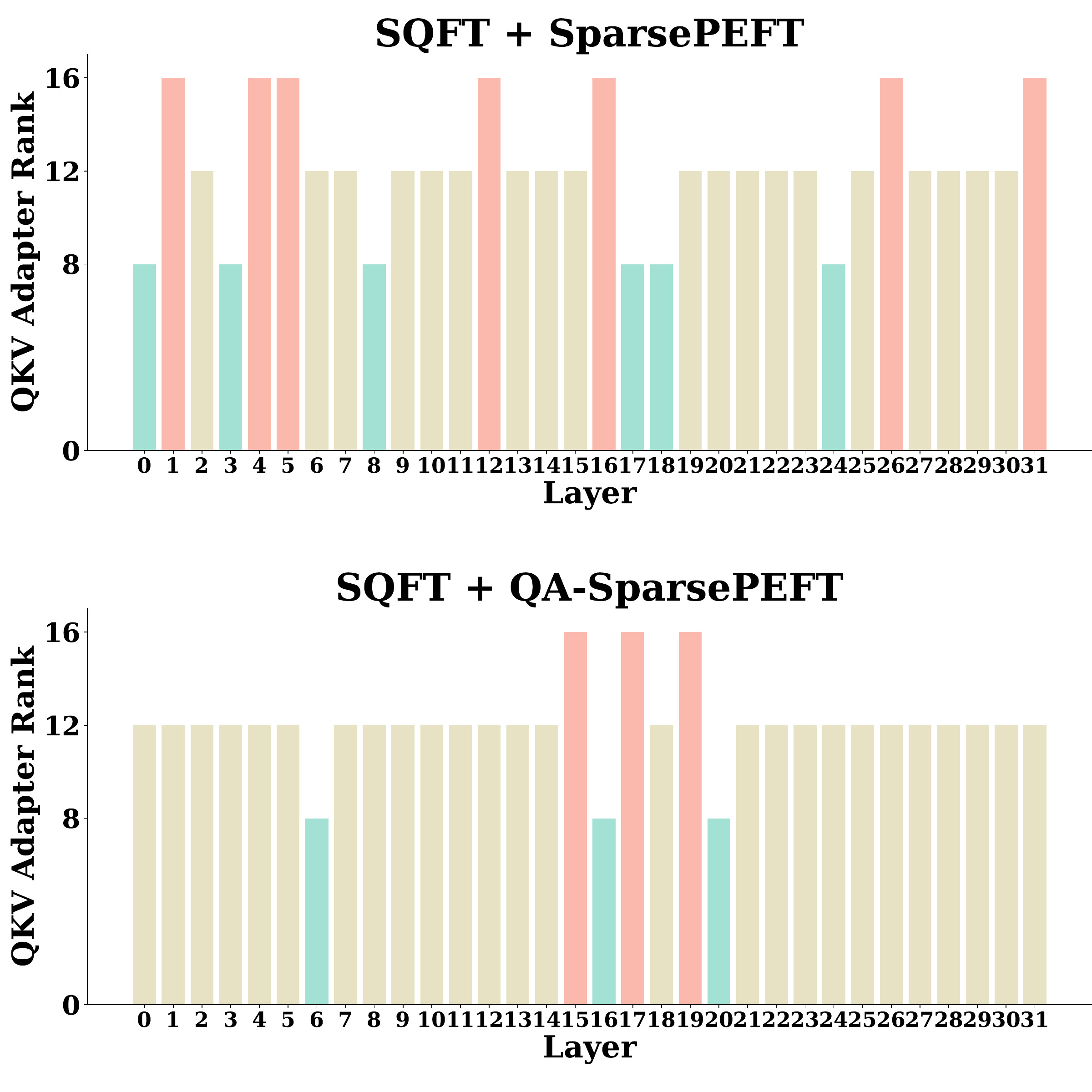}
  \caption{The adapter rank distribution of the optimal configurations obtained from the hill-climbing search algorithm (\textbf{Phi-3-Mini-4K-Instruct} with commonsense reasoning). 
  }
\label{fig:rank_distribution}
\end{figure}

\subsection{Hill-climbing to Better Configurations}

The results presented in the previous sections employ the simple heuristic (as detailed in Section \ref{sec:ex_setup}) to obtain a reference configuration from the NLS search space. However, superior configurations can be discovered with an additional budget. We apply a well-designed hill-climbing search algorithm (Algorithm \ref{alg:hill_climbing} in Appendix), which starts from the configuration derived from the heuristic and explores its neighboring configurations in a hill-climbing matter based on their validation accuracy. 
For this purpose, we employed the validation sets from Arc-e, Arc-c, and OBQA, as other datasets do not provide a validation set.
As demonstrated in Table \ref{tab:hillclimbing}, a more optimal configuration can be discovered, outperforming the default adapter configuration obtained from the heuristic. Exploring further the search space of elastic adapter ranks produces richer adapter distributions as depicted in Figure \ref{fig:rank_distribution}. 
More importantly, the test set results reveal a significant improvement in the performance of the Arc-c and OBQA datasets, which suggests that an appropriate validation set can assist in identifying the optimal adapter configuration.

\begin{figure}
  \centering
  \includegraphics[width=\linewidth]{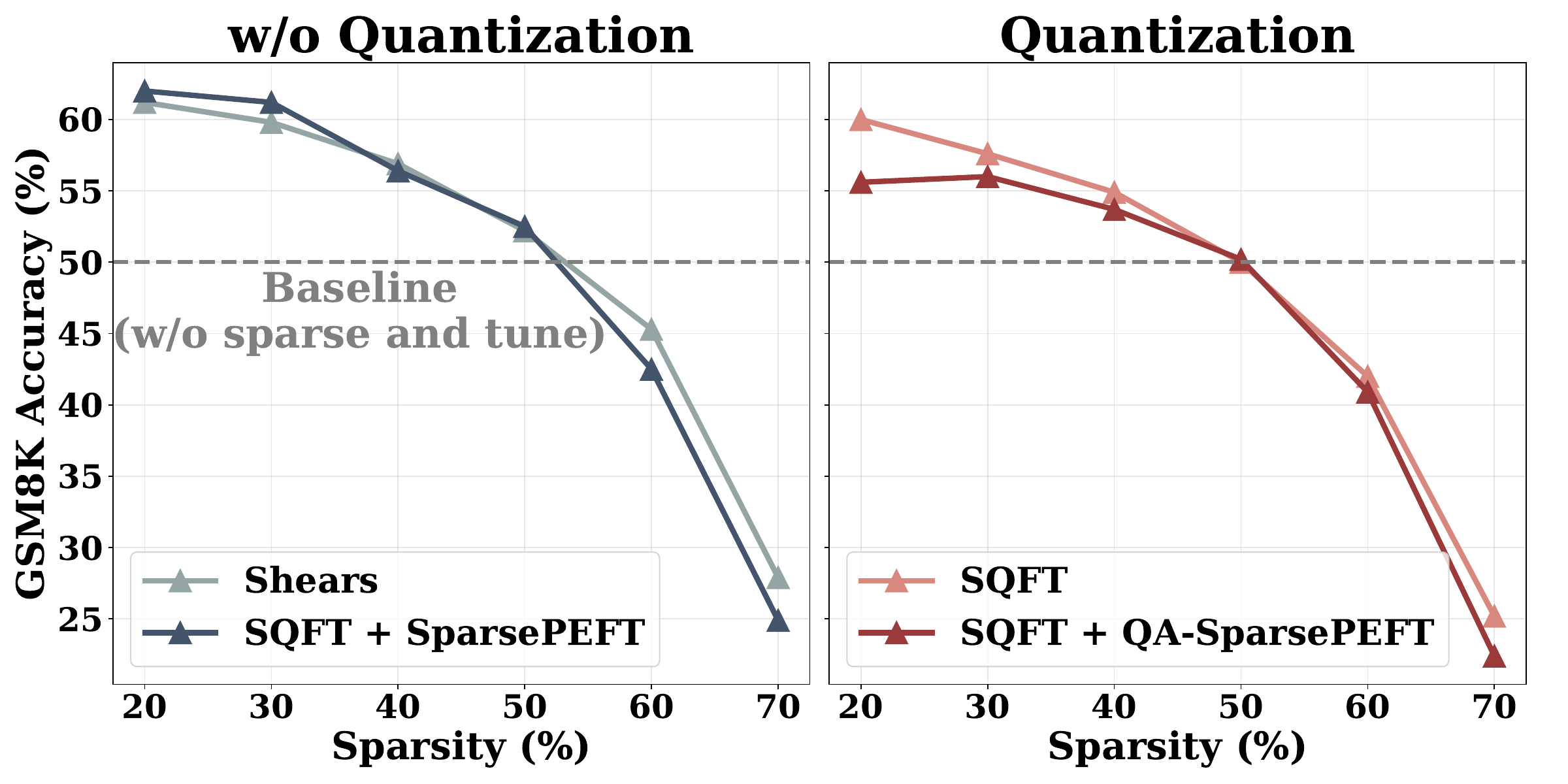}
  \caption{Comparison of various sparsity levels for \textbf{Llama-3-8B} with GSM8K. \proj achieves similar performance as Shears but with the added benefit of merging adapters with different numerical precision.}
\label{fig:gsm8k_llama3_various_sparsity}
\end{figure}

\subsection{Exploring a Broader Range of Sparsity Levels}
\label{sec:broader_range_of_sparsity_levels}

\begin{table*}[hbt]
\setlength{\belowcaptionskip}{0.5pt}
\setlength{\tabcolsep}{6pt}
\caption{Ablation studies for LoRA vs. NLS (\textbf{Llama-3-8B} with GSM8K). Compared to LoRA, NLS obtains significantly better accuracy across all possible pipelines of \proj and different sparsity levels. 
}
\label{tab:gsm8k_llama3_ablation_studies}
\scriptsize 
\centering
\renewcommand\arraystretch{1.1}
\begin{tabular}{lclcccc}
\toprule
\multirow{2}{*}{\textbf{Model}} & \multirow{2}{*}{\textbf{Sparsity}}  & \multirow{2}{*}{\textbf{Method}} & \multirow{2}{*}{\textbf{Mergeable}} & \textbf{Final Precision} & \textbf{Fine-tune}  & \textbf{GSM8K Test}   \\ 
&  & & & \textbf{(Base + Adapter / Base)} &\textbf{Approach}  & \textbf{Accuracy(\%)}   \\
\midrule

\multirow{24}{*}{\textbf{Llama-3-8B}} & \multirow{8}{*}{30\%}& \multirow{2}{*}{Shears}        & \multirow{2}{*}{\xmark} & \multirow{2}{*}{FP16 + FP16} & LoRA  & 58.2  \\
& &         &  &  & NLS & \textbf{59.8$_\text{\textcolor{red}{+1.6}}$} \\
\cdashline{3-7}
& & \multirow{2}{*}{\proj + SparsePEFT (Ours)} & \multirow{2}{*}{\cmark} & \multirow{2}{*}{FP16} & LoRA & 60.0 \\
& &         &  &  & NLS & \textbf{61.2$_\text{\textcolor{red}{+1.2}}$} \\

\cdashline{3-7}

& & \multirow{2}{*}{\proj (Ours)}        & \multirow{2}{*}{\xmark} & \multirow{2}{*}{INT4 + FP16} & LoRA  & 56.7  \\
& &         &  &  & NLS & \textbf{57.6$_\text{\textcolor{red}{+0.9}}$} \\

\cdashline{3-7}

& & \multirow{2}{*}{\proj + QA-SparsePEFT (Ours)} & \multirow{2}{*}{\cmark} & \multirow{2}{*}{INT4} & LoRA  & 54.8 \\
& &         &  &  & NLS & \textbf{56.0$_\text{\textcolor{red}{+1.2}}$} \\

\cline{2-7}

& \multirow{8}{*}{50\%} & \multirow{2}{*}{Shears}        & \multirow{2}{*}{\xmark} & \multirow{2}{*}{FP16 + FP16} & LoRA & 50.6 \\
& &         &  &  & NLS& \textbf{52.2$_\text{\textcolor{red}{+1.6}}$} \\
\cdashline{3-7}
& & \multirow{2}{*}{\proj + SparsePEFT (Ours)} & \multirow{2}{*}{\cmark} & \multirow{2}{*}{FP16} & LoRA  & 50.6 \\
& &         &  &  & NLS & \textbf{52.5$_\text{\textcolor{red}{+1.9}}$} \\
\cdashline{3-7}
& & \multirow{2}{*}{\proj (Ours)}        & \multirow{2}{*}{\xmark} & \multirow{2}{*}{INT4 + FP16} & LoRA  & 48.9  \\
& &         &  &  &NLS & \textbf{50.0$_\text{\textcolor{red}{+1.1}}$} \\
\cdashline{3-7}
& & \multirow{2}{*}{\proj + QA-SparsePEFT (Ours)} & \multirow{2}{*}{\cmark} & \multirow{2}{*}{INT4} & LoRA  & 48.2  \\
& &         &  &  &NLS & \textbf{50.2$_\text{\textcolor{red}{+2.0}}$} \\

\cline{2-7}

& \multirow{8}{*}{70\%}  & \multirow{2}{*}{Shears}        & \multirow{2}{*}{\xmark} & \multirow{2}{*}{FP16 + FP16} & LoRA  & 25.5 \\
& &         &  &  &NLS  & \textbf{27.9$_\text{\textcolor{red}{+2.4}}$} \\
\cdashline{3-7}
& & \multirow{2}{*}{\proj + SparsePEFT (Ours)} & \multirow{2}{*}{\cmark} & \multirow{2}{*}{FP16}  & LoRA & 22.1 \\
& &         &  &  &NLS  & \textbf{24.9$_\text{\textcolor{red}{+2.8}}$} \\
\cdashline{3-7}
& & \multirow{2}{*}{\proj (Ours)}        & \multirow{2}{*}{\xmark} & \multirow{2}{*}{INT4 + FP16} & LoRA  & 24.2   \\
& &         &  &  & NLS & \textbf{25.2$_\text{\textcolor{red}{+1.0}}$} \\
\cdashline{3-7}
& & \multirow{2}{*}{\proj + QA-SparsePEFT (Ours)} & \multirow{2}{*}{\cmark} & \multirow{2}{*}{INT4}  & LoRA & \textbf{22.6$_\text{\textcolor{red}{+0.2}}$}   \\
& &         &  &  & NLS& 22.4 \\

\bottomrule
\end{tabular}
\end{table*}

All our previous experiments employ 50\% sparsity as it is moderate and mild. In this section, we explored the behavior of \proj in a broader range of sparsity levels. 
As shown in Figure \ref{fig:gsm8k_llama3_various_sparsity}, the model’s accuracy experiences a significant drop between a sparsity of 60\% and 70\%. We denote this range as the critical sparsity threshold, representing the boundary at which the model’s performance notably degrades. 
Through our recovery downstream fine-tuning strategy, models with up to 50\% sparsity (even with quantization) can achieve comparable performance with the original dense model (represented by the baseline in the figure) on the downstream task. This 50\% sparsity can be defined as the optimal sparsity level, as it represents the point of balance where the model maintains high performance while achieving computational efficiency. 
Moreover, there is little difference in accuracy between our mergeable and non-mergeable approaches, which illustrates the effectiveness of our proposed SparsePEFT.

\subsection{Cost Analysis of Pipeline Configurations}

The different versions of \proj's pipelines incur various costs that allow users to choose based on their fine-tuning budget. Table \ref{tab:cost_analysis_ranking} details the characteristics of each pipeline configuration, e.g., whether we can merge the adapters, the precision of the based model and the adapters, and the cost of each configuration. 
Two assumptions are made regarding model storage, inference speedup, or memory: merging is better than unmerging due to the overhead from the unmerged adapters, and quantization mode is better than full-precision mode.
As for accuracy, the mergeable method we propose is competitive with the previous non-mergeable method.
Regarding the fine-tuning time, our mergeable method is slightly slower than the non-mergeable method due to the additional mask and adapter calculations.
In summary, \proj with SparsePEFT is the best choice for full-precision mode because it eliminates the adapter's additional path without sacrificing accuracy. 
Suppose memory usage during fine-tuning is a priority for the quantization mode. In that case, vanilla \proj (first configuration in Figure \ref{fig:overall_pipeline}) is the best choice because it only requires the quantized model with little overhead of different precision adapters. Otherwise, \proj with QA-SparsePEFT is better because it can ultimately produce a most efficient model that will be of great benefit at deployment time.

\begin{table}[hbt]
\setlength{\belowcaptionskip}{0.5pt}
\setlength{\tabcolsep}{4pt}
\caption{Cost analysis for different pipelines (\textbf{rank}). ID 1, 2, 3, and 4 represent LoRA/Shears, \proj, \proj + SparsePEFT, and \proj + QA-SparsePEFT, respectively.
}
\label{tab:cost_analysis_ranking}
\scriptsize 
\centering
\renewcommand\arraystretch{1.2}
\begin{tabular}{lcccc}
\toprule
\textbf{ID} & \textbf{1} & \textbf{2} & \textbf{3} & \textbf{4} \\
\midrule
\textbf{Mergeable} &  \xmark  &  \xmark  &  \cmark  &  \cmark \\
\textbf{Final Precision} & FP16 + FP16 & INT4 + FP16 & FP16 & INT4\\
\midrule
\textbf{Model Storage ($\downarrow$)} & \multicolumn{4}{c}{$1 > 3 \gg 2 > 4$} \\
\textbf{Fine-tuning Time ($\downarrow$)} & \multicolumn{4}{c}{$1 \approx 2 < 3 \approx 4$} \\
\textbf{Fine-tuning Memory ($\downarrow$)} & \multicolumn{4}{c}{$2 < 1 \approx 3 \approx 4$} \\
\textbf{Inference Speedup ($\uparrow$)} & \multicolumn{4}{c}{$4 > 2 > 3 > 1$} \\
\textbf{Inference Memory ($\downarrow$)} & \multicolumn{4}{c}{$4 < 2 < 3 < 1$} \\
\textbf{Accuracy ($\uparrow$)} & \multicolumn{4}{c}{$1 \approx 3 > 2 \approx 4$} \\
\bottomrule
\end{tabular}
\end{table}

\begin{table}[hbt]
\setlength{\belowcaptionskip}{0.5pt}
\setlength{\tabcolsep}{3pt}
\caption{Cost analysis for different pipelines (\textbf{value}). ID 1, 2, 3, and 4 represent LoRA/Shears, \proj, \proj + SparsePEFT, and \proj + QA-SparsePEFT, respectively. All numbers are tested on a single Tesla V100-SXM2-32GB GPU. Both training and inference are conducted on \textbf{Llama-3-8B} with GSM8K, with a batch size of 16 during training. 
}
\label{tab:cost_analysis_value}
\scriptsize  
\centering
\renewcommand\arraystretch{1.2}
\begin{tabular}{lcccc}
\toprule
\textbf{ID} & \textbf{1} & \textbf{2} & \textbf{3} & \textbf{4} \\
\midrule
\textbf{Mergeable} &  \xmark  &  \xmark &  \cmark    &  \cmark \\
\textbf{Final Precision} & FP16 + FP16 & INT4 + FP16 & FP16 & INT4\\
\midrule
\textbf{Model Storage} & 16.33 GB & 6.00 GB & 16.07 GB  & 5.74 GB \\
\textbf{Fine-tuning Speed } & \multirow{2}{*}{0.3} & \multirow{2}{*}{0.3} & \multirow{2}{*}{0.2}  & \multirow{2}{*}{0.2} \\
\textbf{(steps per second)} &  \\
\textbf{Fine-tuning Memory} & 30 GiB & 21 GiB & 30 GiB  & 30 GiB \\
\bottomrule
\end{tabular}
\end{table}

\subsection{Ablation Studies - LoRA vs NLS}
\label{sec:ablation_Studies}

As shown in Table \ref{tab:gsm8k_llama3_ablation_studies}, the ablation studies across 30\%, 50\%, and 70\% sparsity highlight the benefits of elastic adapters and the Neural Low-rank Adapter Search (NLS), which enhance the performance of the models fine-tuned by \proj. Compared to vanilla LoRA, SQFT with SparsePEFT and NLS further reduces the accuracy gap to the dense or non-quantized models. We include more results with additional sparsity levels in the Appendix, which show the benefits of using SQFT with NLS for sparse and quantized models.

%% file: content/5_conclusion.tex
\section{Conclusion}
\label{sec:conclusion}

Large pre-trained models often require fine-tuning to downstream target tasks and compression to utilize them in resource-constrained environments. This paper presents \proj, a low-cost fine-tuning solution for low precision and sparse foundation models. \proj solves challenges when merging sparse (and quantized) base models and dense (with different numerical precision) adapters without losing the induced sparsity in the base model while delivering high-performing fine-tuned models. \proj's models and code are available at \href{https://github.com/IntelLabs/Hardware-Aware-Automated-Machine-Learning}{https://github.com/IntelLabs/Hardware-Aware-Automated-Machine-Learning}.

\section*{Limitations and Ethical Considerations}

Large pre-trained models have gained popularity and are the base of many applications. However, these models are often used indiscriminately with little analysis of their potential failures and consequences. \proj solely focuses on these large models' efficient fine-tuning and compression. However, users of \proj should also consider the limitations of these models before deployment in environments where they can cause harm or conflict. Although compressing and fine-tuning these models on a particular downstream task would make them perform better, more studies are needed regarding the effects of this specialization.

We demonstrate \proj on several pre-trained models. The benefits obtained from the proposed solution might transfer smoothly to other transformer-based models. However, there might also be models and datasets in which additional considerations must be taken. For instance, in our current experiments, we have noticed that in the case of OpenELM-1.1B \cite{openelm}, fine-tuning on math reasoning datasets, e.g., GSM8K, does not result in high accuracy, and more experimentation is needed. There is also the case in which a pre-trained model might have been trained on a particular benchmark, a form of data contamination, which is difficult to confirm since often the details of the training data are not shared publicly \cite{zhang2024careful}. In these cases, inducing sparsity might result in a drop in accuracy on that particular benchmark.  

Due to the many unknowns and complexity of current large models, it is essential to take measures to prevent their use in sensitive applications. With insights obtained by the research community in the years to come, understanding the intricacies of these models will help us use them beneficially and safely. 

\section*{Acknowledgments}

We are grateful to Michael Beale from Intel Labs, who helped us set up the infrastructure for sharing our models during the review stage and the final release and guided us through open-sourcing our compressed models. We also thank the anonymous reviewers for their insightful suggestions, which helped us improve the paper.

%% file: content/appendix.tex
\section*{Appendix}
\section{Related Work}
\label{sec:related}

Generative pre-trained models, often based on the Transformer architecture \cite{Transformer_NIPS2017}, require the application of compression techniques to reduce their significant computational cost and to address challenges, e.g., related to memory bandwidth. Classic compression techniques like pruning and quantization have been adapted for the age of LPMs, removing inefficiencies that cannot be tolerated when dealing with billions of parameters. We discuss them in more detail next. 

\paragraph{Pruning} Inducing sparsity, either by zeroing out weights or activations or removing network elements, can improve the efficiency of LPMs during inference, provided that they are executed on a runtime that can exploit sparse patterns. Pruning has a long history \cite{LeCunBrainDamage}, but with the advent of LPMs, traditional methods\cite{hoeflerSparsity21}, e.g., Magnitude Pruning \cite{HAGIWARA1994207}, have been replaced by new approaches that are suited for the challenges of these models with their large number of parameters. SparseGPT \cite{frantar-sparsegpt} proposes a one-shot pruning method for transformer-based models that trade minimal accuracy drop for increasing sparsity levels. The method approaches LPMs' pruning layer-wise with an efficient weight reconstruction algorithm that incrementally prunes the weight matrix elements. Wanda \cite{sun2023wanda} proposes a more straightforward approach that does not require weight updates, computing a score using the weight magnitude and the norm of input activations. This approach obtains better results than SparseGPT. Recently, BESA \cite{xu2024besa} has improved over SparseGPT and Wanda by targeting individual transformer blocks and allocating sparsity per layer using a differentiable method. 
These approaches induce sparsity on pre-trained models and are evaluated on zero-shot benchmarks. Our end-to-end solution, \proj, focuses on further adapting the sparsified models to new tasks or datasets. 

\paragraph{Quantization} 
With the advent of large pre-trained foundation/frontier models (LPMs), quantization approaches have evolved to address the challenges of scale and memory bandwidth. Due to the high cost of retraining these models to recover accuracy degradation, special consideration has to be taken when incorporating compression techniques, like quantization-aware training in foundation models. Post-training, one-shot quantization methods have prevailed, obtaining quantized versions of large models in hours. 
LLM.Int8() was among the first Int8 quantization procedures for large-scale transformer-based PLMs
\cite{dettmers2022LLMInt8}. Using vector-wise quantization and mixed-precision decomposition, LLM.Int8() demonstrated that it can effectively confront the outliers that emerge in activations, which makes traditional quantization methods fail in models with more than 6.7B parameters. In a contemporary work, after running thousands of experiments with various large pre-trained models, it was demonstrated that 4-bit parameters can reach optimal performance compared to other bit-precisions in the 3 to 16-bit range \cite{case4bit22}. 
ZeroQuant \cite{NEURIPS20ZeroQuant} quantizes GPT-3 models, obtaining a reduction in latency up to 4.16x by utilizing group-wise quantization for weights, token-wise quantization for activations, and layer-by-layer knowledge distillation. SmoothQuant \cite{xiao2023smoothquant} makes activations easier to quantize by smoothing them and compensating this operation with a transformation of the weights, resulting in improved results over ZeroQuant and LLM.Int8(). GPTQ is another good representative of one-shot quantization approaches designed especially for LPMs \cite{frantar-gptq}. GPTQ builds on the learnings from Optimal Brain Quantization (OBQ) \cite{frantar2022obc} and applies layer-wise quantization to the full-precision weights of a base LPM. We incorporate GPTQ as the default quantization method in \proj's pre-fine-tuning stage.

\paragraph{Parameter-efficient Fine-tuning (PEFT)} Due to their large number of parameters, it is too costly to fine-tune pre-trained large models. Updating all their weights to improve their performance in a downstream task might require devices with large memory capacity. PEFT techniques attempt to address this challenge by avoiding the update of all weights in the pre-trained model. For instance, low-rank (LoRA) adapters \cite{hu2022lora} use a fraction (often less than 1\%) of additional weights to adapt the model to a new task. LoRA adapters, $\boldsymbol{B}$ and $\boldsymbol{A}$, are utilized to reparameterize a linear projection, $\boldsymbol{Y} = \boldsymbol{WX}$, keeping the weights, $\boldsymbol{W}$, frozen and updating only the low-rank adapter matrices, $\boldsymbol{A}$ and $\boldsymbol{B}$, i.e., $\boldsymbol{Y} = \boldsymbol{WX} + \boldsymbol{BAX}$. 

\begin{algorithm*}
\label{alg:hill}
  \small
  \caption{Hill-climbing Subnetwork Search}
  \label{alg:hill_climbing}
  \begin{algorithmic}[1]
    \Require{Number of turns $T$, Number of neighbors $N$, Neighbor step size $S$, Number of evaluation samples $M$, Heuristic configuration $c_h$, Validation dataset $\mathcal{D}$}
    \Ensure{Optimal configuration $c^*$}
    \State $c_a \gets c_h$  \Comment{\textit{Initialize anchor with the heuristic configuration}}
    \State $V \gets \{c_h\}$ \Comment{\textit{Initialize the set of visited configurations}}
    \State $\mathcal{D}_M \gets $ Sample($\mathcal{D}$, $M$) \Comment{\textit{Create a proxy dataset by randomly sampling $M$ samples from $\mathcal{D}$}}
    \For{$t \gets 1$ to $T$}
      \State $\mathcal{C} \gets $ Neighbor-sample($c_a$, $N$, $S$) - $\mathcal{V}$ \Comment{\textit{Sample $N$ unvisited $S$-step neighbor configs}}
      \State $\mathcal{V} \gets \mathcal{V} \cup \mathcal{C}$ \Comment{\textit{Add the sampled configurations to the set of visited configurations}}
        \State $c_m \gets $ MaxAcc(Eval($\mathcal{D}_M$, $\mathcal{C}$))  \Comment{\textit{The config with the maximum accuracy on proxy data}}
        \If{$Acc(c_m) > Acc(c^*)$}
          \State $c_a \gets c_m$ \Comment{\textit{Update anchor configuration if the new configuration has higher accuracy}}
        \EndIf
    \EndFor
  \State $c^* \gets c_a$   \Comment{\textit{The optimal configuration is the final anchor configuration}}
  \State \Return $c^*$
  \end{algorithmic}
\end{algorithm*}

Recently, Shears proposed Neural Low-rank Adapter Search \cite{shears} and demonstrated that LoRA adapters can be made elastic to allow for the application of weight-sharing schemes and keeping the original weights of the model frozen and compressed, e.g., inducing sparsity before the fine-tuning stage. However, a challenge that emerges is that merging the dense adapters with the sparse weights results in the overall loss of sparsity. LoRAPrune has attempted to address this challenge by using the weights and gradients of the LoRA adapters to remove elements in the model's weights  \cite{zhang2023loraprune}. As demonstrated in the main sections of the paper, \proj proposes an alternative method for merging the dense adapters with a minimal drop in accuracy.

\section{Hyperparameters}

The hyperparameters used in our main experiments are shown in Table \ref{tab:hyperparameters}.

\begin{table*}[hbt]
\setlength{\belowcaptionskip}{5pt}
\setlength{\tabcolsep}{3pt}
\caption{Hyperparameters used in our experiments. For all approaches with NLS, we explored several manually designed search spaces and identified the optimal configuration for each pipeline. Note that in our experiments involving GSM8K and math instruction tuning, we conducted trials over 3 or 4 epochs and reported the best results achieved. Interestingly, \proj with QA-SparsePEFT often necessitates extended training periods to exploit its quantization-aware capabilities fully.}
\scriptsize
\centering
\renewcommand\arraystretch{1.2}
\begin{tabular}{lcclcccccc}
\toprule
\multirow{2}{*}{\textbf{Model}} & \multirow{2}{*}{\textbf{Task}} & \multirow{2}{*}{\textbf{Sparsity}} & \multirow{2}{*}{\textbf{Method}} & \multirow{2}{*}{\textbf{Epoch}}  & \textbf{Batch} & \textbf{Learning} &  \multirow{2}{*}{\textbf{Adapter rank}} & \textbf{Adapter} & \textbf{Adapter} \\ 
 &  &  &  &  & \textbf{size} & \textbf{rate} &  & \textbf{alpha} & \textbf{target modules} \\ 
\midrule
\multirow{6}{*}{\textbf{Llama-3-8B}} & \multirow{6}{*}{GSM8K} &  \multirow{6}{*}{50\%} & LoRA & 3 & 16 & 3e-4 & 32 & 64 & Q, K, V, Up, Down \\
 &  &  & Shears & 3 & 16 & 3e-4 & 32,28,24,20,16 & 64 & Q, K, V, Up, Down \\
 &  &  & \proj + SparsePEFT & 3 & 16 & 3e-4 & 48,32,16 & 64 & Q, K, V, Up, Down \\
 &  &  & GPTQ + LoRA & 3 & 16 & 3e-4 & 32 & 64 & Q, K, V, Up, Down \\
 &  &  & \proj & 3 & 16 & 3e-4 & 40,32,24 & 64 & Q, K, V, Up, Down \\
 &  &  & \proj + QA-SparsePEFT & 4 & 16 & 3e-4 & 48,32,16 & 64 & Q, K, V, Up, Down \\
\midrule
\multirow{6}{*}{\textbf{Mistral-7B-v0.3}} & \multirow{6}{*}{GSM8K} &  \multirow{6}{*}{50\%} & LoRA & 3 & 16 & 3e-4 & 32 & 64 & Q, K, V, Up, Down \\
 &  &  & Shears & 3 & 16 & 3e-4 & 32,28,24,20,16 & 64 & Q, K, V, Up, Down \\
 &  &  & \proj + SparsePEFT & 3 & 16 & 3e-4 & 32,28,24,20,16 & 64 & Q, K, V, Up, Down \\
 &  &  & GPTQ + LoRA & 3 & 16 & 3e-4 & 32 & 64 & Q, K, V, Up, Down \\
 &  &  & \proj & 3 & 16 & 3e-4 & 32,28,24,20,16 & 64 & Q, K, V, Up, Down \\
 &  &  & \proj + QA-SparsePEFT & 4 & 16 & 3e-4 & 32,28,24,20,16 & 64 & Q, K, V, Up, Down \\
\midrule
\multirow{6}{*}{\textbf{Mistral-7B-v0.3}} & \multirow{6}{*}{Math} &  \multirow{6}{*}{50\%} & LoRA & 3 & 16 & 3e-4 & 32 & 64 & Q, K, V, Up, Down \\
 &  &  & Shears & 3 & 16 & 3e-4 & 32,28,24,20,16 & 64 & Q, K, V, Up, Down \\
 &  &  & \proj + SparsePEFT & 3 & 16 & 3e-4 & 32,28,24,20,16 & 64 & Q, K, V, Up, Down \\
 &  &  & GPTQ + LoRA & 3 & 16 & 3e-4 & 32 & 64 & Q, K, V, Up, Down \\
 &  &  & \proj & 3 & 16 & 3e-4 & 32,28,24,20,16 & 64 & Q, K, V, Up, Down \\
 &  &  & \proj + QA-SparsePEFT & 4 & 16 & 3e-4 & 32,28,24,20,16 & 64 & Q, K, V, Up, Down \\
\midrule

\multirow{6}{*}{\textbf{Phi-3-Mini-4K-Instruct}} & \multirow{6}{*}{Math} & \multirow{6}{*}{50\%}  & LoRA & 3 & 16 & 3e-4 & 32 & 64 & Qkv \\
 &  &  & Shears & 3 & 16 & 3e-4 & 48,40,32,24,16 & 64 & Qkv \\
 &  &  & \proj + SparsePEFT & 3 & 16 & 3e-4 & 48,32,16 & 64 & Qkv \\
 &  & & GPTQ + LoRA & 3 & 16 & 3e-4 & 32 & 64 & Qkv \\
 &  &  & \proj & 3 & 16 & 3e-4 & 32,28,24,20,16 & 64 & Qkv \\
 &  &  & \proj + QA-SparsePEFT & 3 & 16 & 3e-4 & 48,32,16 & 64 & Qkv \\
\midrule
\multirow{6}{*}{\textbf{Phi-3-Mini-4K-Instruct}} & \multirow{6}{*}{CS} &\multirow{6}{*}{50\%}  & LoRA & 3 & 16 & 1e-4 & 16 & 32 & Qkv \\
 &  & & Shears & 3 & 16 & 1e-4 & 16,12,8 & 32 & Qkv \\
 &  &  & \proj + SparsePEFT & 3 & 16 & 1e-4 & 16,12,8 & 32 & Qkv \\
 &  &  & GPTQ + LoRA & 3 & 16 & 1e-4 & 16 & 32 & Qkv \\
 &  &  & \proj & 3 & 16 & 1e-4 & 16,12,8 & 32 & Qkv \\
 &  &  & \proj + QA-SparsePEFT & 3 & 16 & 1e-4 & 16,12,8 & 32 & Qkv \\
\bottomrule
\end{tabular}
\label{tab:hyperparameters}
\end{table*}

\section{Hill-climbing search algorithm}

We propose Algorithm \ref{alg:hill_climbing} to start from the reference configuration (Section \ref{sec:ex_setup}) and systematically explore its neighbors. Table \ref{tab:hillclimbing} in the main paper shows the benefits of using any available budget to execute this algorithm and discover better-performing models. 

\section{Additional Sparsity Levels and Ablation Studies for Llama-3 on GSM8K}
\label{sec:additional_sparsity_level}

We conducted additional experiments and ablations studies with different sparsity levels and compared the underlying NLS approach to LoRA. Table \ref{tab:gsm8k_llama3_ablation_studies_various_sparsity} shows that up to high sparsity levels, \proj delivers high-performing models. 

\begin{table*}[hbt]
\setlength{\belowcaptionskip}{0.5pt}
\setlength{\tabcolsep}{6pt}
\caption{Ablation studies for various sparsity levels (\textbf{Llama-3-8B} with GSM8K). 
}
\label{tab:gsm8k_llama3_ablation_studies_various_sparsity}
\scriptsize 
\centering
\renewcommand\arraystretch{1}
\begin{tabular}{lclcccc}
\toprule
\multirow{2}{*}{\textbf{Model}} & \multirow{2}{*}{\textbf{Sparsity}}  & \multirow{2}{*}{\textbf{Method}} & \multirow{2}{*}{\textbf{Mergeable}} & \textbf{Final Precision} & \textbf{Fine-tune}  & \textbf{GSM8K Test} \\ 
&  & & & \textbf{(Base + Adapter / Base)} &\textbf{Approach}  & \textbf{Accuracy(\%)} \\
\midrule

\multirow{73}{*}{\textbf{Llama-3-8B}} & 0\%  & w/o tune &  -  & FP16 & -  & 50.0   \\

\cline{2-7}

& \multirow{12}{*}{20\%} &  \multicolumn{5}{c}{\textit{w/o Quantization}} \\
\cdashline{3-7}
& & w/o tune &  -  & FP16 & - & 47.5  \\

& & \multirow{2}{*}{Shears}        & \multirow{2}{*}{\xmark} & \multirow{2}{*}{FP16 + FP16} & LoRA & 58.7 \\
& &         &  &  & NLS& \textbf{61.2$_\text{\textcolor{red}{+2.5}}$} \\

& & \multirow{2}{*}{\proj + SparsePEFT} & \multirow{2}{*}{\cmark} & \multirow{2}{*}{FP16} & LoRA  & 60.3 \\
& &         &  &  & NLS & \textbf{62.0$_\text{\textcolor{red}{+1.7}}$} \\
\cdashline{3-7}
& & \multicolumn{5}{c}{\textit{Quantization}} \\
\cdashline{3-7}
& & w/o tune &  -  & INT4 & - & 36.6  \\

& & \multirow{2}{*}{\proj}        & \multirow{2}{*}{\xmark} & \multirow{2}{*}{INT4 + FP16} & LoRA  & 57.8  \\
& &         &  &  &NLS & \textbf{60.0$_\text{\textcolor{red}{+2.2}}$} \\

& & \multirow{2}{*}{\proj + QA-SparsePEFT} & \multirow{2}{*}{\cmark} & \multirow{2}{*}{INT4} & LoRA  & 54.7    \\
& &         &  &  &NLS & \textbf{55.6$_\text{\textcolor{red}{+0.9}}$} \\

\cline{2-7}

& \multirow{12}{*}{30\%}  &  \multicolumn{5}{c}{\textit{w/o Quantization}} \\
\cdashline{3-7}
& & w/o tune &  -  & FP16 & -  & 40.9  \\

& & \multirow{2}{*}{Shears}        & \multirow{2}{*}{\xmark} & \multirow{2}{*}{FP16 + FP16} & LoRA  & 58.2  \\
& &         &  &  & NLS & \textbf{59.8$_\text{\textcolor{red}{+1.6}}$} \\

& & \multirow{2}{*}{\proj + SparsePEFT} & \multirow{2}{*}{\cmark} & \multirow{2}{*}{FP16} & LoRA & 60.0  \\
& &         &  &  & NLS & \textbf{61.2$_\text{\textcolor{red}{+1.2}}$} \\

\cdashline{3-7}

& & \multicolumn{5}{c}{\textit{Quantization}} \\

\cdashline{3-7}

& & w/o tune &  -  & INT4 & - & 30.3   \\

& & \multirow{2}{*}{\proj}        & \multirow{2}{*}{\xmark} & \multirow{2}{*}{INT4 + FP16} & LoRA  & 56.7  \\
& &         &  &  & NLS & \textbf{57.6$_\text{\textcolor{red}{+0.9}}$} \\

& & \multirow{2}{*}{\proj + QA-SparsePEFT} & \multirow{2}{*}{\cmark} & \multirow{2}{*}{INT4} & LoRA  & 54.8 \\
& &         &  &  & NLS & \textbf{56.0$_\text{\textcolor{red}{+1.2}}$} \\

\cline{2-7}

& \multirow{12}{*}{40\%} &  \multicolumn{5}{c}{\textit{w/o Quantization}} \\
\cdashline{3-7}
& & w/o tune &  -  & FP16 & - & 31.6  \\

& & \multirow{2}{*}{Shears}        & \multirow{2}{*}{\xmark} & \multirow{2}{*}{FP16 + FP16} & LoRA & 56.9  \\
& &         &  &  & NLS & 56.9 \\

& & \multirow{2}{*}{\proj + SparsePEFT} & \multirow{2}{*}{\cmark} & \multirow{2}{*}{FP16} & LoRA  & \textbf{57.9$_\text{\textcolor{red}{+1.5}}$}    \\
& &         &  &  & NLS & 56.4 \\
\cdashline{3-7}
& & \multicolumn{5}{c}{\textit{Quantization}} \\
\cdashline{3-7}
& & w/o tune &  -  & INT4 & - & 20.1   \\

& & \multirow{2}{*}{\proj}        & \multirow{2}{*}{\xmark} & \multirow{2}{*}{INT4 + FP16} & LoRA  & 54.9  \\
& &         &  &  &NLS & 54.9 \\

& & \multirow{2}{*}{\proj + QA-SparsePEFT} & \multirow{2}{*}{\cmark} & \multirow{2}{*}{INT4} & LoRA  & 53.4   \\
& &         &  &  &NLS & \textbf{53.7$_\text{\textcolor{red}{+0.3}}$} \\

\cline{2-7}

& \multirow{12}{*}{50\%} &  \multicolumn{5}{c}{\textit{w/o Quantization}} \\
\cdashline{3-7}
& & w/o tune &  -  & FP16 & - & 12.5  \\

& & \multirow{2}{*}{Shears}        & \multirow{2}{*}{\xmark} & \multirow{2}{*}{FP16 + FP16} & LoRA & 50.6\\
& &         &  &  & NLS& \textbf{52.2$_\text{\textcolor{red}{+1.6}}$} \\

& & \multirow{2}{*}{\proj + SparsePEFT} & \multirow{2}{*}{\cmark} & \multirow{2}{*}{FP16} & LoRA  & 50.6 \\
& &         &  &  & NLS & \textbf{52.5$_\text{\textcolor{red}{+1.9}}$} \\
\cdashline{3-7}
& & \multicolumn{5}{c}{\textit{Quantization}} \\
\cdashline{3-7}
& & w/o tune &  -  & INT4 & - & 7.0   \\

& & \multirow{2}{*}{\proj}        & \multirow{2}{*}{\xmark} & \multirow{2}{*}{INT4 + FP16} & LoRA  & 48.9  \\
& &         &  &  &NLS & \textbf{50.0$_\text{\textcolor{red}{+1.1}}$} \\

& & \multirow{2}{*}{\proj + QA-SparsePEFT} & \multirow{2}{*}{\cmark} & \multirow{2}{*}{INT4} & LoRA  & 48.2  \\
& &         &  &  &NLS & \textbf{50.2$_\text{\textcolor{red}{+2.0}}$} \\

\cline{2-7}

& \multirow{12}{*}{60\%} &  \multicolumn{5}{c}{\textit{w/o Quantization}} \\
\cdashline{3-7}
& & w/o tune &  -  & FP16 & - & -  \\

& & \multirow{2}{*}{Shears}        & \multirow{2}{*}{\xmark} & \multirow{2}{*}{FP16 + FP16} & LoRA & 39.9   \\
& &         &  &  & NLS& \textbf{45.3$_\text{\textcolor{red}{+5.4}}$} \\

& & \multirow{2}{*}{\proj + SparsePEFT} & \multirow{2}{*}{\cmark} & \multirow{2}{*}{FP16} & LoRA  & 40.7 \\
& &         &  &  & NLS & \textbf{42.5$_\text{\textcolor{red}{+1.8}}$} \\
\cdashline{3-7}
& & \multicolumn{5}{c}{\textit{Quantization}} \\
\cdashline{3-7}
& & w/o tune &  -  & INT4 & - & -   \\

& & \multirow{2}{*}{\proj}        & \multirow{2}{*}{\xmark} & \multirow{2}{*}{INT4 + FP16} & LoRA  & 40.1  \\
& &         &  &  &NLS & \textbf{42.0$_\text{\textcolor{red}{+1.9}}$} \\

& & \multirow{2}{*}{\proj + QA-SparsePEFT} & \multirow{2}{*}{\cmark} & \multirow{2}{*}{INT4} & LoRA  & 37.6    \\
& &         &  &  &NLS & \textbf{40.9$_\text{\textcolor{red}{+3.3}}$} \\

\cline{2-7}

& \multirow{12}{*}{70\%} &  \multicolumn{5}{c}{\textit{w/o Quantization}} \\
\cdashline{3-7}
& & w/o tune &  -  & FP16 & - & -   \\

& & \multirow{2}{*}{Shears}        & \multirow{2}{*}{\xmark} & \multirow{2}{*}{FP16 + FP16} & LoRA  & 25.5 \\
& &         &  &  &NLS  & \textbf{27.9$_\text{\textcolor{red}{+2.4}}$} \\

& & \multirow{2}{*}{\proj + SparsePEFT} & \multirow{2}{*}{\cmark} & \multirow{2}{*}{FP16}  & LoRA & 22.1 \\
& &         &  &  &NLS  & \textbf{24.9$_\text{\textcolor{red}{+2.8}}$} \\
\cdashline{3-7}
& & \multicolumn{5}{c}{\textit{Quantization}} \\
\cdashline{3-7}
& & w/o tune &  -  &INT4 & - & -  \\

& & \multirow{2}{*}{\proj}        & \multirow{2}{*}{\xmark} & \multirow{2}{*}{INT4 + FP16} & LoRA  & 24.2  \\
& &         &  &  & NLS & \textbf{25.2$_\text{\textcolor{red}{+1.0}}$} \\

& & \multirow{2}{*}{\proj + QA-SparsePEFT} & \multirow{2}{*}{\cmark} & \multirow{2}{*}{INT4}  & LoRA & \textbf{22.6$_\text{\textcolor{red}{+0.2}}$}  \\
& &         &  &  & NLS& 22.4 \\

\bottomrule
\end{tabular}
\end{table*}

\section{How does \proj perform without sparsity?}

\begin{table*}[hbt]
\setlength{\belowcaptionskip}{0.5pt}
\setlength{\tabcolsep}{10pt}
\caption{Results from adapting \textbf{Llama-3-8B} to GSM8K without introducing sparsity.
}
\label{tab:gsm8k_main}
\scriptsize 
\centering
\renewcommand\arraystretch{1.2}
\begin{tabular}{llcccc}
\toprule
\multirow{2}{*}{\textbf{Model}}  & \multirow{2}{*}{\textbf{Method}} & \multirow{2}{*}{\textbf{Mergeable}} & \textbf{Final Precision}  & \textbf{Fine-tune} & \textbf{GSM8K Test} \\ 
&  & & \textbf{(Base + Adapter / Base)} &\textbf{Approach}  & \textbf{Accuracy(\%)}   \\
\midrule

\multirow{6}{*}{\textbf{Llama-3-8B}}   & w/o tune &  -  & FP16  & - & 50.0  \\
\cline{2-6}
& \multicolumn{5}{c}{\textit{Quantization}} \\
\cdashline{2-6}
& w/o tune &  -  & \textbf{INT4}  & - & 36.6   \\
& GPTQ + LoRA        & \xmark & INT4 + FP16 & LoRA & 58.8  \\
& \proj (Ours)       & \xmark & INT4 + FP16  & NLS & \textbf{61.0}\\
& \multirow{2}{*}{\proj + QA-SparsePEFT (Ours)} & \multirow{2}{*}{\cmark} & \multirow{2}{*}{\textbf{INT4}}  & LoRA & 55.8  \\
&  &  &  & NLS  & 57.2  \\

\bottomrule
\end{tabular}
\end{table*}

In the main paper, we explored \proj with both sparse + non-quantized and sparse + quantized settings. However, we are also interested in what happens to \proj if there is no sparsity. Here, we investigate \proj's performance with quantization alone. As shown in Table~\ref{tab:gsm8k_main}, without sparsity, the quantized model reduces the accuracy from 50\% to 36.6\%. With the help of fine-tuning, the baseline GPTQ + LoRA improves accuracy to 58.8\%. At the same time, our \proj method further enhances performance, achieving 61.0\% accuracy with NLS fine-tuning, demonstrating that NLS outperforms LoRA in the non-sparse setting.
However, for \proj + QA-SparsePEFT, while NLS outperforms LoRA, the accuracy is slightly lower compared to \proj. The advantage is that it results in an INT4 model. In summary, users must balance accuracy and efficiency based on their requirements to choose the optimal approach.